\documentclass{article}
\usepackage{ijcai21}

\usepackage{times}
\usepackage{soul}
\usepackage{url}
\usepackage[hidelinks]{hyperref}
\usepackage[utf8]{inputenc}
\usepackage[small]{caption}
\usepackage{graphicx}
\usepackage{amsmath}
\usepackage{amsthm}
\usepackage{booktabs}
\usepackage{algorithm}
\usepackage{algorithmic}
\title{Continual Distributed Learning for Crisis Management}

\author{
Aman Priyanshu$^1$\footnote{Contact Author}\and
Mudit Sinha$^2$\And
Shreyans Mehta$^3$
\affiliations
$^1$Manipal Institute of Technology\\
$^2$University of Massachusetts Amherst\\
$^3$K J Somaiya College of Engineering\\
\emails
amanpriyanshusms2001@gmail.com
}

\begin{document}

\maketitle

\begin{abstract}
  Social media platforms such as Twitter, Facebook etc can be utilised as an important source of information during disaster events. This information can be used for disaster response and crisis management if processed accurately and quickly. However, the data present in such situations is ever-changing, and using considerable resources during such crisis is not feasible. Therefore, we have to develop a low resource and continually learning system that incorporates text classification models which are robust against noisy and unordered data. We utilised Distributed learning which enabled us to learn on resource-constrained devices, then to alleviate catastrophic forgetting in our target neural networks we utilized regularization. We then applied federated averaging for distributed learning and to aggregate the central model for continual learning.
\end{abstract}

\section{Introduction}

Social media platforms play an essential role in today’s era. Connecting over 300 million active users, Twitter is one of the largest platforms for micro-blogging.Platforms such as Linkedln,Twitter and Facebook have become an essential part of today’s society, constructing close-knit online communities and creating safe spaces for young voices. These platforms can be used for rapid knowledge dissemination \cite{AKM:1} and are therefore an excellent source for information dispersal during crises.

Crisis management requires a rapid and precise response by support and rescue services. On social media, information is constantly relayed by an active user base during disaster events. This can provide crucial insights and breakthroughs during disaster events. For example, creating a missing persons report will be easier with our model.These days, even support and safety operations are put up on social media sites, cataloguing them will help both the people affected by disasters and the public services working to aid them.   Twitter and other such platforms become mediums for rapid communication during disasters and relay a plethora of vital information. It is a largely community driven project as it depends on people offering support and relaying information.  The aggregate information conveyed every second during such events can be used for assistance or rapid response. Other media such as televisions, new broadcasts, digital newspapers have yet to tap into this source properly, they do use it sometimes but their methods are inefficient and smaller in scale compared to a trained model which organizes a large pool of data and is built for the purpose of helping authorities and people during disasters specifically. 

Although these sites provide a continuous flow of information, most of it remains unordered and therefore untappable. To enable the use of this information for public services, it must be organized and catalogued. However, manual aggregation through a volunteering process \cite{KANKANAMGE2019101097} is time-consuming and may not be rapid enough to enable efficient utilization of information received. Therefore, we aim to provide an automated system for classifying data which enables quick response. We use raw textual data from twitter for developing a continually learning text classification tool, which can aid public utility service in times of crisis.

A significant concern with crisis management algorithms is the redundant nature of trained models. As time progresses, there is rapid development in the situation, and only a continual learning model may be capable of learning across such rapid changing tasks. At the same time, heavy resource requirements are not feasible during a crisis event and therefore, a volunteer backed distributed learning algorithm may be a more viable solution. In this paper, we explore the concepts of distributed learning coupled with continual learning for crisis management. We further support our claims by running experiments on our proposed methodology.

\section{Literature Review}

\subsection{Continual Learning}

Continual or lifelong learning refers to the ability to continually learn over time by accommodating new knowledge while retaining previously learned experiences \cite{PARISI201954}. The research in this field has found significant development by utilising the concept of regularisation. Joint learning was one of the first few successful implementations of continual learning, it required interleaving samples from each task \cite{Caruana1997}. However, this methodology becomes increasingly cumbersome as the number of tasks increases or the types of samples increase. It also requires memory from previous tasks and data samples, making it more resource constraining.

This led to the development of Learning without Forgetting (LwF) \cite{li2017learning}, which required only newer samples for learning. Although similar to joint training, the method does not require old data or reference points; instead, it uses regularisation to compensate for the network forgetting an entire sequence of old data.

\begin{equation}\label{equ:1}
    \begin{array}{l}
        {Loss\ =\ \lambda}_0\mathcal{L}_{old}\left(Y_0,{\hat{Y}}_0\right) \\
        +\mathcal{L}_{new}\left(Y_n,{\hat{Y}}_n\right)+R\left({\hat{\Theta}}_s,{\hat{\Theta}}_0,{\hat{\Theta}}_{new}\right) 
    \end{array}
\end{equation}

By using Equation \ref{equ:1} the model is able to retain the previous performance, learnt on the old tasks as well as continually learn and update for newer tasks.

\subsection{Federated Averaging}

Konečný et al. proposed Federated Learning as a framework for distributed learning while securing data privacy \cite{konecny2016federated}. Over the last few years, numerous Federated Learning frameworks and architectures have been proposed for various problem statements, including computer vision \cite{10.5555/3196160.3196245} and next-word prediction in mobile keyboards \cite{kairouz2021advances}. It offers a practical solution to the drawbacks of distributed learning. Federated learning provides a convenient workaround and offers to solve collaborative learning problems optimally. One of the most efficient and straightforward means of deploying a Federated learning model is the Federated averaging algorithm. It locally updates client-weights and communicates the updated weights to a central server, which acts as an aggregator. Once aggregated here, the central server calculates the geometric mean over the received weights from different clients \cite{mcmahan2017communicationefficient}. Once the central model is trained over multiple local epochs and multiple aggregation steps, it achieves a significant enough performance for it to be deployed. We used this implementation of federated learning in our proposed methodology.

The combination of two independent branches can bring newfound implementation in various tasks. Here, we utilise both continual as well as distributed learning for information classification during a crisis. We empirically show that our methodology is robust against noisy data and is able to give a significant performance across its continual deployment.

\section{Methodology}

Recent advances in the field of distributed and continual learning have contributed vastly to the field of artificial intelligence. Our aim with this paper is to bring together these two technologies to provide immediate technological support for crisis management. Our proposed methodology employs Federated Averaging to classify tweets across different targets from different volunteers. The aim is to produce an efficient model that can be trained in a distributed fashion while employing some of the most substantial resources available in Natural Language Processing, namely, sentence encoders.

Unlike word embedders which are used to derive sentence inference, sentence encoders generally return a single embedding for the entire sentence. This allows us to take these embedding as an input for Local Client models. For federated model training, the weight updates are communicated to the central server at the end of the local training. This allows for faster convergence of the federated learning model. 

During local training, we employ regularisation to prevent the network from forgetting as the model progresses through new tasks. This leads to a continual learning process allowing the model to update to the latest training sets.

\begin{figure}[h]
\centerline{\includegraphics[scale=0.6]{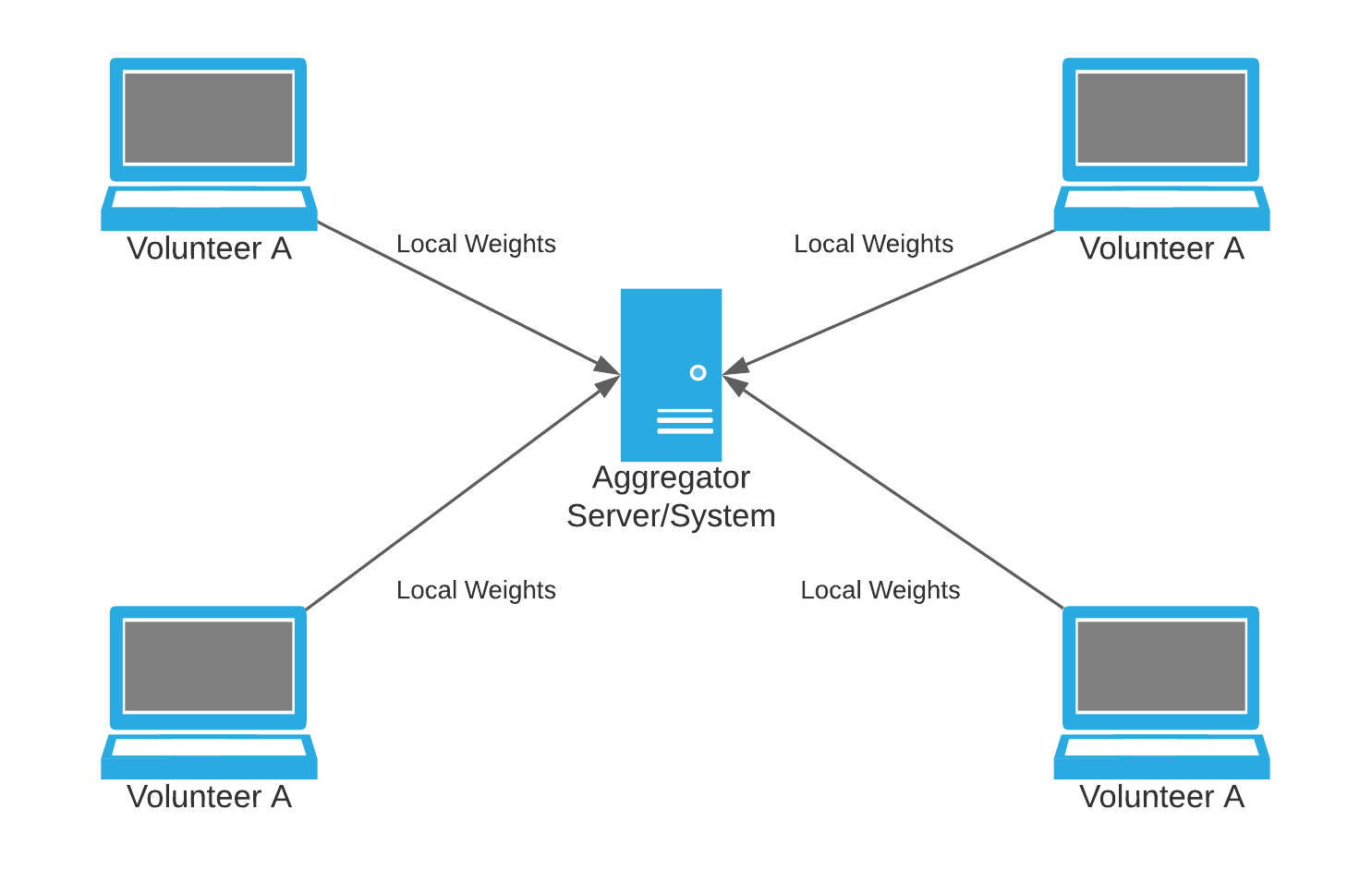}}
\caption{Aggregation Step: Displays ideal procedure for aggregation, where multiple volunteers return their updated local models to the central server.}
\label{fig}
\end{figure}

The algorithm begins by initializing a central model with trainable parameters or weights, which is then distributed across all client devices (here volunteer systems). Given the central server's initial model parameters, each device accesses its local data, consisting of the accumulated tweets. This data is then fed into a settled-on sentence encoder, which is used for feature extraction. This input data is then divided into fixed batches, and backpropagation is applied to each of them. For this step, we utilize a special regularized loss function. After iterating over all batches, the updated parameter values are sent to the central server for aggregation, and the whole process is repeated.

\begin{figure}[h]
\centerline{\includegraphics[scale=0.5]{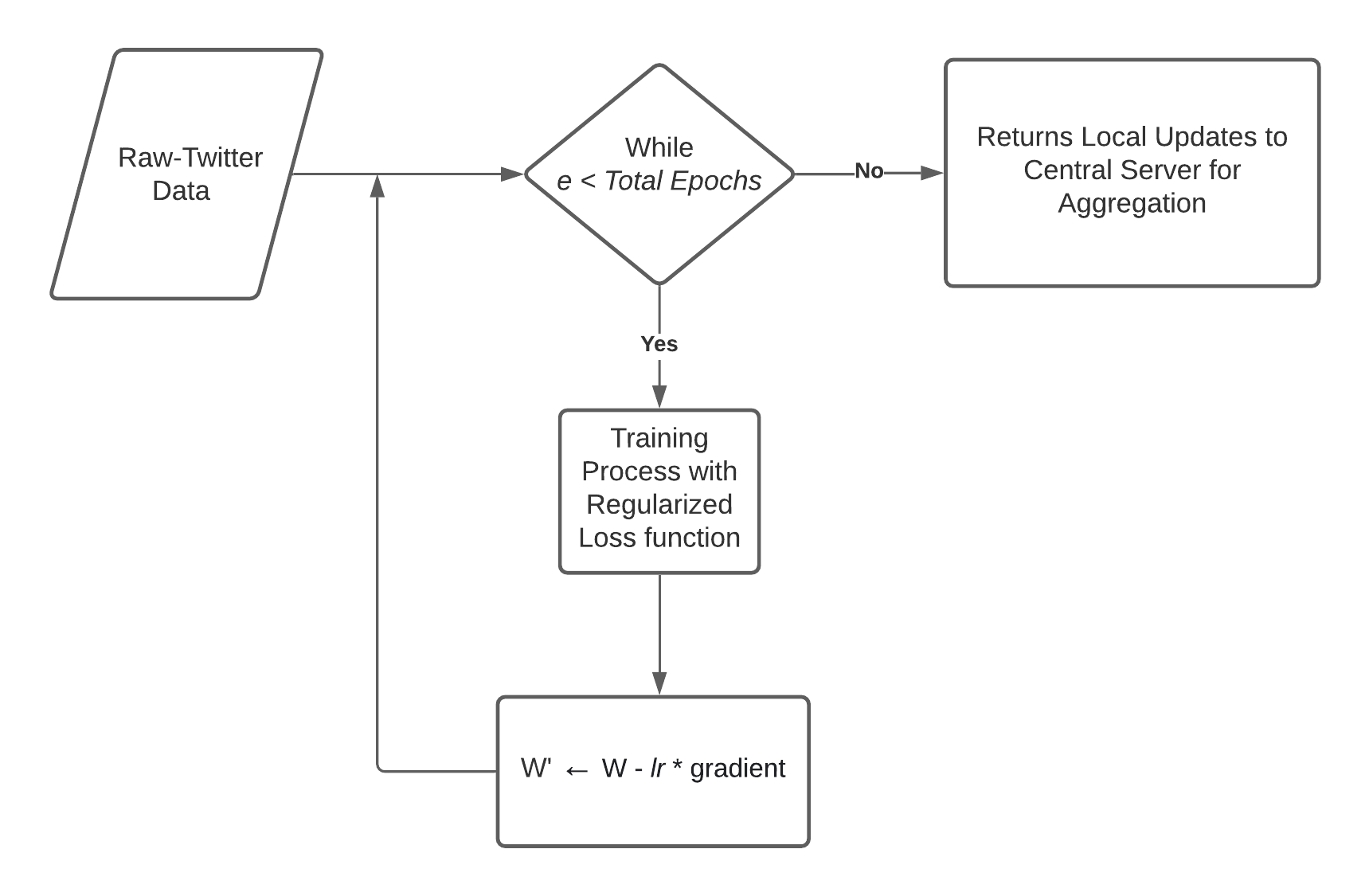}}
\caption{Local Training: The models here are trained using locally available data, thereby making it computationally inexpensive. The loss function used here is defined in Equation \ref{equ:1}}
\label{fig}
\end{figure}

The integration of LwF(learning without forgetting) and federated averaging algorithm  helps us to achieve our aim of producing a continual distributed learning model. We now study the results and performance of our proposed methodology.

\section{Experiments}

\subsection{Implementation Details}

\begin{table*}[ht]
\centering
\begin{tabular}{|c|c|c|}
\hline
\textbf{Baselines} & \textbf{Training Accuracy} & \textbf{Test Accuracy}\\
\hline
Neural Network (Neither Federated Learning nor Continual Learning) & 82.057 & 78.201 \\
With Continual Learning only & 84.283 & 76.211 \\
With Federated Learning only & 79.157 & 63.924 \\
\hline
\end{tabular}
\caption{\label{baselines}
The table displays the baseline results as developed across different algorithms.
}
\end{table*}

Our proposed methodology uses sentence embeddings to classify disaster-related tweets for aiding crisis management and recovery. The sentence embeddings allow sentences to be embedded in a context-aware manner, allowing any sequential classifiers to label our data correctly. We employ universal sentence encoder lite \cite{cer2018universal} for extracting our features. This encoding model makes use of a deep averaging network (DAN) and then the features are passed through a feedforward deep neural network (DNN) to produce sentence embeddings. This deep neural network then maps the given sentences to a 512-dimensional vector.

Sentence encoders show significant improvements for contextual and intent-based classification of sentences, thereby increasing the performance of the textual classifiers. The features are extracted from Universal Sentence Encoders (USE) and then fed into a different fully connected neural architectures for learning.

Parameters in the model are optimized with respect to the binary cross-entropy loss coupled with regularization for lifelong learning as given by Equation \ref{equ:1}.

Our analysis is aimed at finding the appropriate client distribution , while maintaining performance. As our model is trained in a distributed or federated fashion, we experimented with different number of clients, to find an adequate trade-off between model performance and number of clients. Since, the number of clients is inversely proportional to the computational intensity required for training, by finding the best trade-off we achieved the most optimal parameters. Then we employed our methodology for different network architectures, which allowed us to have more generalized results. 

\subsection{Dataset}

The HumAID Twitter dataset consists of thousands of manually annotated tweets that have been collected during 19 major natural disaster events including earthquakes, hurricanes, wildfires, and floods, which occurred across the World between 2016 and 2019. \cite{humaid2020}. For our analysis, we will be taking the 3 recent events, occurring between 2018 and 2019. The events are Cyclone Idai, Greece Wildfires and the Maryland Floods, which are each a different type of crisis. Utilizing the continual learning capacity of our methodology, we wish to display its learning capacity. The dataset used consists of over 6,000 tweets. We used the following distribution which was specified in the dataset ;631 tweets for validations set, 1229 tweets for testing and 4332 tweets for training.The tweets are distributed in 10 labels as given below , 
\begin{itemize}
  \item caution and advice
  \item displaced people and evacuations
  \item infrastructure and utility damage
  \item injured or dead people
  \item missing or found people
  \item not humanitarian
  \item other relevant information
  \item requests or urgent needs
  \item rescue volunteering or donation effort
  \item sympathy and support
\end{itemize}

\subsection{System Details}

Our experiments were executed on a central server, the training process was implemented using the Federated Averaging algorithm. The client devices, were simulated on a single system, each simulated client device was trained parallelly, independent of each other. The system consisted of a  device with 8GB RAM and GeForce GTX 1650, 4GB GPU. The proposed methodology is implemented using TensorFlow. The Adam optimizer updates the model parameters, with an initial learning rate of 0.001 \cite{kingma2017adam}. The batch size is fixed to 32 for all networks.

\begin{figure*}[h]
\centerline{\includegraphics[scale=0.3]{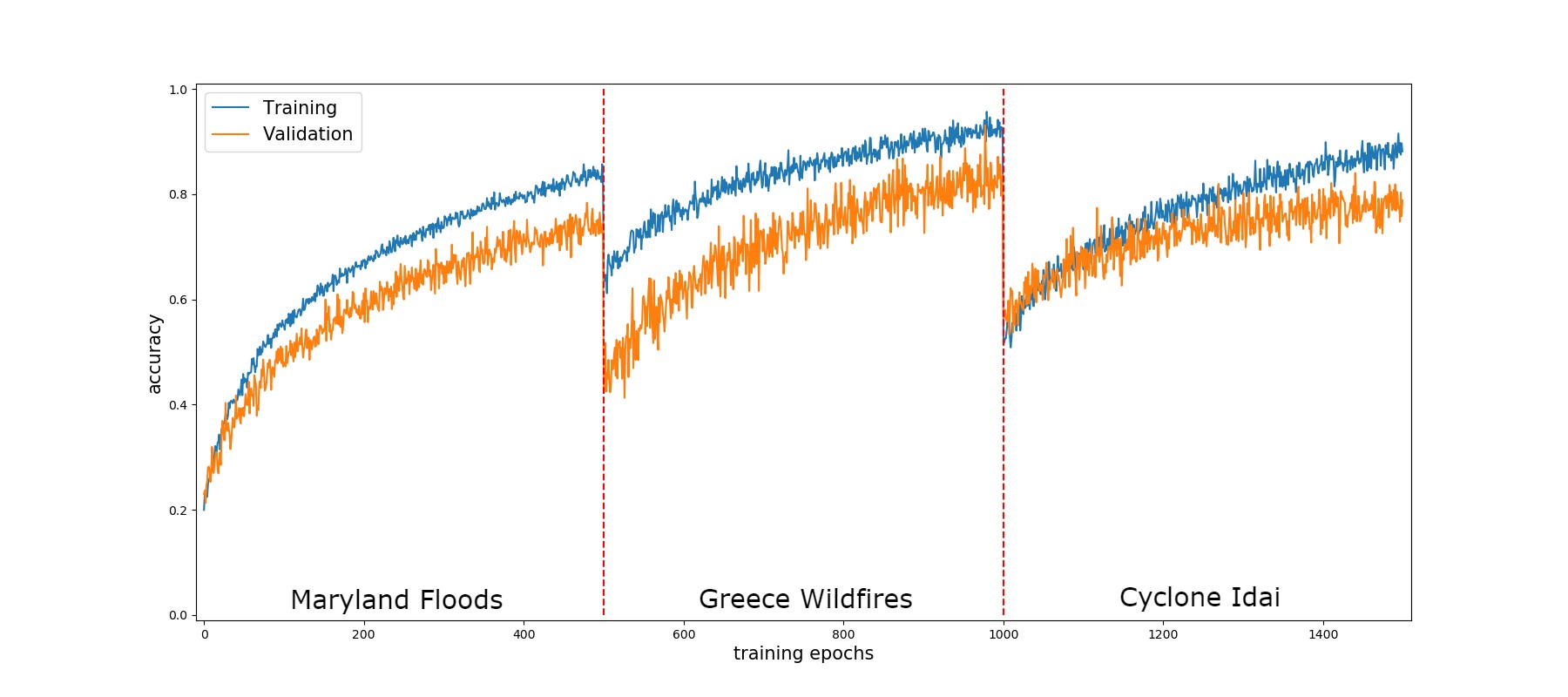}}
\caption{The figure, shows the training progression across different catastrophic events. The above image conveys the continual learning property of our proposed methodology. We see the network is quicker to learn, once it has been trained on previous tasks.}
\label{progression}
\end{figure*}

\section{Results}

The experiments were conducted keeping two major points in mind.
\begin{enumerate}
    \item Model Complexity to Reduce Computational Overhead.
    \item Number of Clients to Performance Trade-off.
\end{enumerate}

In our experiments we validate and compare multiple hyper-parameters. For comparison we use baseline results as shown in Table~\ref{baselines}. For our proposed methodology, we begin our analysis by evaluating models of different depth (number of layers) while keeping number of neurons constant. We defined models of $depth\ \in\ [3,\ 5,\ 10,\ 15,\ 25]$, which allowed us to have a varied sample size to query from, for evaluation. The models were trained in a centralized fashion, using the Adam optimizer. We ran each model for $50 epochs$ and display their results in Table \ref{depth-results}.

\begin{table}[h]
\centering
\begin{tabular}{lll}
\hline
\textbf{Depth} & \textbf{Test Loss} & \textbf{Test Accuracy}\\
\hline
3 & 1.8979 & 71.5969 \\
5 & 1.8726 & 72.7740 \\
10 & 1.1226 & 80.5270 \\
15 & 1.2255 & 78.8404 \\
25 & 1.5781 & 74.62337 \\
\hline
\end{tabular}
\caption{\label{depth-results}
The table displayed the results achieved by centralized baseline models on the entire dataset. We took the distribution available within the HumAID Twitter dataset for validation, training and testing allowing us to verify our models accurately.}
\end{table}

We can see from our experimental results, that we are able to train a generalized network with a depth of 10 layers, each activated by the ReLU activation function which yields the most optimal results. Here, we have defined each layer with 100 neurons each. We finally feed the extracted features to the Softmax function layer, where it is computed into the probabilities of our labels. We can see from Table \ref{depth-results} that the model learns uptil $depth = 10$ after which it begins to overfit on the training set.

Having concluded the optimal depth and size of our network, we now begin describing the relation between number of clients and performance of our model. The federated averaging algorithm may suffer from performance loss, if data becomes Non-IID, i.e., Non-Independent and Identically Distributed Data, this may happen due to smaller data size. Therefore we look at Twitter's API which allows 15,000 tweets to be scraped in a window of 15 minutes. Hence, we only pick number of samples per client between 500 and 1500. $samples\_per\_client\ \in\ [500, 1500]$. Since, we have only 4,332 tweets in our training set, we range our analysis between $[3, 9]$ clients.

As we've seen with a distributed learning system, reducing the number of divisions or distributions improves the neural network's performance. Our experiments show likewise, in our analysis presented in Table~\ref{number-of-clients-results}. These results are evaluated over 500 epochs per event, and their testing and training performance is measured by the cumulative mean of all three events. 

\begin{table}[h]
\centering
\begin{tabular}{p{1.7cm}p{1.75cm}p{1.6cm}}
\hline
\textbf{Number of Clients} & \textbf{Training Accuracy} & \textbf{Test Accuracy}\\
\hline
3 & 86.493 & 77.265 \\
5 & 84.147 & 74.932 \\
7 & 83.033 & 70.033 \\
9 & 81.485 & 66.362 \\
\hline
\end{tabular}
\caption{\label{number-of-clients-results}
The table displays the trade-off between number of clients and network performance.
}
\end{table}

The results show a distinctive relation between client sample size and model overfitting. We can see, as the number of clients increases, sample size decreases which results in network overfitting. Therefore, fewer albeit distributed number of clients may results in better deployment. Two methods to combat such overfitting is:

\begin{enumerate}
    \item Transfer Learning
    \item Increasing Local Data Size.
\end{enumerate}

The training progression over the course of three crisis, i.e., Cyclone Idai, GreeceWildfires and the Maryland Floods are displayed in Figure \ref{progression}

We can clearly see the advantage of continual learning, as it is easier for the network to adapt to the task at hand. The adaptability of the network can be attributed to is regularized loss function. The proposed methodology proves robust against change in events and can be a viable option for rapid learning during crisis management. The continual learning can be seen having an advantage, as displayed in Table~\ref{depth-results} (Accuracy: 80.52\%), where we achieve better results than baseline results in Table~\ref{baselines} (Accuracy: 78.201\%).

\section{Conclusion}

Our proposed methodology for Continual Distributed Learning for Crisis Management offers a convenient and optimal solution towards disaster management. The procedure employs continual or life long learning with consistent results throughout three distinct kinds of disasters. We also utilise the distributed learning prowess of the federated averaging algorithm to train models in a resource constraint setting. The system as a whole can give sustainable results for deployment purposes during very dire situations. Looking at some of the labels, "displaced people and evacuations", "infrastructure and utility damage", "injured or dead people", and "missing or found people", we can ascertain that these will help public support services rescue operations. These labels are an essential part of crisis management and recovery, and therefore we included them within our training set.
Our work and analysis gives a novel approach towards crisis management, which can be used for live tweet analysis to extract crucial information. The use of continual learning allows the model to learn with higher performance. Federated learning as a form of distributed learning also reduces the load on single systems, it can also be used to bypass Twitter API's limited scraping capacity, thereby leveraging technology to achieve greater heights. We hope to further improve our algorithm by including different types of continual learning algorithms and further experimenting with scalability to enhance communications for distributed deployment.

\bibliographystyle{named}
\bibliography{ijcai21}

\end{document}